\documentstyle[11pt,twocolumn,named,a4wide,program,psfig]{article}

\renewcommand{\baselinestretch}{1.03} 

\date{May 1994}
\title{\protect{\large \bf {\renewcommand{\baselinestretch}{0.9} 
EMPHATIC GENERATION: EMPLOYING THE THEORY OF\\[.3ex] SEMANTIC 
EMPHASIS FOR TEXT GENERATION\\[-.3ex]}}}
\author{\bf Elke Teich, Beate Firzlaff, John A. Bateman%
\thanks{Also on indefinite leave from the Penman Project,
USC/Information Sciences Institute, Marina del Rey, Los Angeles.}\\[0.05in]
{\small GMD/Institut f\"ur Integrierte Publikations-
und Informationssysteme}\\
{\small Dolivostra\ss e 15, D-64293 Darmstadt, Germany}\\[0.05in]
{\small {\em e-mail:} \{teich,firzlaff,bateman\}@darmstadt.gmd.de}}

\begin{document}
\thispagestyle{empty}
\maketitle
\thispagestyle{empty}

{\sl
Paper presented at COLING '94 (Kyoto, Japan) but 
not to be found in the printed proceedings.

Cmp-lg No: cmp-lg/9704012.}

\begin{center} {\bf Abstract} \end{center}
\vspace*{-2ex}
{\renewcommand{\baselinestretch}{0.8} \small
The paper deals with the problem of text generation and planning
approaches making only limited formally specifiable contact with
accounts of grammar. We propose an enhancement of a systemically-based
generation architecture for German \cite{Bateman-etal91-penang} by
aspects of the theory of semantic emphasis \cite{Kunze91}.  Doing
this, we gain more control over both concept selection in generation
and choice of fine-grained grammatical variation.}

\section{\protect{\large \bf INTRODUCTION}}

The extension of linguistic representation to levels of abstraction
above syntax is an important theoretical goal; current efforts in this
direction include \cite{Alshawi92,Grover-etal93-alvey,Jackendoff90}.
However, problematic with most of these developments is their
restriction to, as it is termed in systemic-functional theory
\cite{Halliday78}, ideational information. As \cite{Grover-etal93-alvey}
report, diverse sentences such as {\em He found it in the park}  and
{\em It was in the park that he found it} are assigned identical
semantic representations by the Alvey grammar, and it is common for
variations such as these to be relegated to `pragmatic'
interpretations of invariant semantic forms. We claim that such
variation also requires a semantic representation based on a {\em textual} semantics
that augments the existing ideational semantics. The importance of such a
broadening of semantic representations is already clear in work on
text generation \cite{horacekzock93,Meteer91,Bateman-etal93-enlgw},
and has been argued also for analysis
\cite{Matthiessen-etal91-kyushu}. Unfortunately, accounts of text
organization and text planning achieved within text generation often
make only limited formally specifiable contact with accounts of
grammar
(e.g.,~\cite{GroszSidner86,MannThompson87,Hovy87-thesis,Hovy-etal92-nlgw});
and contrariwise, formal accounts such as discourse representation
theory, although beginning to make contact with higher levels of
rhetorical organization (e.g.,~\cite{LascaridesAsher91}), are
typically restricted to describing anaphoric relations and
quantification.  In this paper, we present one particular extension to
a textual semantics, showing its integration and use in dealing with
some related problems in text generation.
The extension is based on the  semantic  analysis  of  the
structure   of   lexical   fields   in   German   developed   by   Kunze
(e.g.,~\cite{Kunze91}), combining  this  with  the systemic-functionally
driven mode of text  generation pursued in  the text generation  project
{\sc     komet}~\cite{Bateman-etal91-penang}.       A      computational
representation of  the  semantic  information  posited  by  Kunze's
theory of semantic emphasis  ({\em Theorie der  semantischen Emphase\/})
 is under  development for  the
lexical    entries    of    the     text    analysis    project     {\sc
kontext}~\cite{FirzlaffHaenelt92}. We describe here how this information is now used
as an  additional  source  of  functional constraints during grammatical
decision making and how this allows a natural contact with certain text organization
decisions.   We briefly  illustrate the  work with  two examples: first,
our approach to  a central  problem in  knowledge-based natural language
processing, that of how to relate domain models to levels of  linguistic
knowledge and  processing;  and  second,  a  demonstration of the co-constraints
between emphasis distribution and certain textual decisions.
Finally, 
we discuss the directions that this work now opens
up for  future  investigation,  including  application of NLP components
to real-world domains \cite{Teich-etal94-anlp} and 
generalizations  to languages other than German (cf. \cite{Kunze92}).

\section{\protect{\large \bf EMPHASIS THEORY}}

\label{emphase}

The theory of semantic emphasis~\cite{Kunze91} proposes explanations
concerning the meaning of situation descriptions communicated by natural
language texts and its relationship to possible syntactic realizations.
One aspect of the theory will be outlined here, namely how a syntactic
realization depends on semantics. Moreover, it will be shown how
grammatical features can be derived systematically.

The theory provides prototypical descriptions of situations. These
descriptions are called {\em basic semantic schemes\/}. They
are given in terms of predicate-argument-structures called
{\em propositions\/}. For instance, the basic semantic scheme for
situations of 
change-of-possession is: 

\noindent
{\small
\begin{tabular}[t]{@{}l@{}l@{}}
    ( cause ( & act (a)\\
              & et ( \begin{tabular}[t]{@{}l}
                         bec ( \begin{tabular}[t]{@{}l@{}l@{}}
                                   have ( a1, a2 ))\\ 
                               \end{tabular} \\
                         bec ( not ( \begin{tabular}[t]{@{}l@{}l@{}}
                                         have ( a3, a4 ))))))\\
                                     \end{tabular} 
                     \end{tabular}
\end{tabular}}

This can be paraphrased as: An action of {\em a\/} causes {\em a1\/} to get {\em
a2\/} and {\em a3\/} to lose {\em a4\/}.\footnote{The
variables are to be filled in by the names of the referents.} Since
this description is prototypical it provides just one transferred
object: it is denoted by {\em a2\/} and {\em a4\/} because it can be
regarded from different points of view. Furthermore, {\em ref(a)\/} might either be the same as {\em ref(a1)\/} or as {\em ref(a3)\/}, 
but {\em ref(a1)\/} and {\em ref(a3)\/} must be different.

So, each participant of a situation may be referred to more than once
in the corresponding description. Each of these references corresponds
to a specific role (deep case) which, in turn, mirrors a specific
point of view towards the referent. The roles are derived
systematically rather than being determined in a more or less
intuitive way (e.g., \cite{Fillmore68}): they are derived according to a set of well-defined recursive rules (cf.
\cite[pp78--89]{Kunze91}); the derivation process follows the
propositional structure bottom-up. A {\em basic predicate\/} has at least one
{\em elementary\/} argument (represented by some variable) to which an initial role value is
assigned. The other predicates only take {\em propositional\/} arguments
and modify the (initial or intermediate) role values assigned to the
elementary arguments. For instance, the basic predicate 
{\em have\/} assigns the role {\em \verb|<|locat, have\verb|>|\/} to its first argument as initial value. The predicate {\em bec\/} further specifies {\em \verb|<|locat,
have\verb|>|\/} as {\em \verb|<|goal,
have\verb|>|\/}. The predicate {\em et\/}, on the other hand, never changes a
role. The predicate {\em cause\/} does not affect 
a {\em \verb|<|goal,
have\verb|>|\/} in its second argument.

The roles derived for the basic semantic scheme in question are:

\begin{center}
{\small \begin{tabular}[t]{@{}l@{}l@{}}
a \/ & \verb|<|agens, act\verb|>|\\
a1 \/ & \verb|<|goal, have\verb|>|\\
a2 \/ & \verb|<|to-obj, have\verb|>|\\ 
a3 \/ & \verb|<|source, have\verb|>|\\
a4 \/ & \verb|<|from-obj, have\verb|>|\\
\end{tabular}}
\end{center}

The second column of this table is the {\em maximum case frame\/} of
verbs that can be used to describe a change-of-possession in which one
object is transferred. However, in a phrase that describes a situation only some roles of the maximum case frame are
verbalized. Roles that are not verbalized are said to be {\em blocked\/}.

Moreover, some aspect of a situation is put into the
foreground, which means that in a suitable phrase, the corresponding role(s) is (are)
verbalized with {\em semantic emphasis\/}. In terms of the theory of semantic emphasis this is reflected 
by the {\em parameter of emphasis\/}, which is assigned 
to partial propositions of a basic semantic scheme. These assignments are the result of a rule-based 
distribution. A basic semantic scheme entails the information where to start the 
distribution. As far as the change-of-possession is concerned 
the starting point is the second argument of the predicate {\em cause\/}, 
namely the proposition with the predicate {\em et\/}. Therefore the {\em
et\/}-proposition has emphasis. A proposition that has emphasis
distributes it top-down to one of its arguments. Accordingly a
proposition that has emphasis can only be the argument of a
proposition that also has emphasis. Consequently, one of the
{\em have\/}-propositions has emphasis, and the {\em
act\/}-proposition may have emphasis.

According to a general rule at least one of the roles of a proposition
with emphasis must not be blocked, which means that it has to be
verbalized. Furthermore, its grammatical realization must be in
nominative, genitive, dative or accusative case. The choice of the
grammatical case mainly depends on the role. Secondly, it is
determined by the subset of roles that are not blocked and belong to
propositions with emphasis. On the other hand, the roles of
propositions without emphasis need not be verbalized at all, but if
one of them is verbalized, its grammatical realization can only be by
oblique case, i.e. by a prepositional object. The choice of suitable
prepositions depends on the role.\footnote{The theory of semantic
emphasis has been worked out for German. However, most of its
principles apply to other languages as well.}

In Figure~\ref{exs} we present some sample sentences. Their
propositional descriptions are derived from the basic semantic scheme
of change-of-possession. Note that if we add emphasis information and
select the roles that are to be verbalized, we construct the {\em
semantic forms} derivable from the basic semantic scheme. (In the
figure, ``---'' indicates that the corresponding role is blocked,
i.e. no grammatical case is assigned to it. Propositions with basic
predicates that have emphasis and the corresponding information
concerning the grammatical realizations are in bold face.)

\begin{figure*}
\rule{\textwidth}{0.2mm}
(1) Sie verliert den Schl\"ussel. (She loses the key.)

{\footnotesize
\begin{tabular}[t]{@{}l@{}l@{}}
    ( cause ( & act ( \begin{tabular}[t]{@{}l@{}l@{}}
                          \verb|<|agens, act\verb|>|: ref(she) 
                          \/ & \dots \verb|[| --- \verb|]| ) \\
                      \end{tabular}\\
              & et ( \begin{tabular}[t]{@{}l}
                         bec ( have ( \begin{tabular}[t]{@{}l@{}l}
                                               \verb|<|goal, have\verb|>|: a1, 
                                               \/ & \dots \verb|[| --- \verb|]|\\ 
                                               \verb|<|to-obj, have\verb|>|: ref(key) 
                                               \/ & \dots \verb|[| --- \verb|]| )) \\
                                           \end{tabular}\\
                         bec ( not ( {\bf have ( \begin{tabular}[t]{@{}l@{}l}
                                                \verb|<|source, have\verb|>|: ref(she), 
                                                \/ & \dots \verb|[| nominative \verb|]| \\
					        \verb|<|from-obj, have\verb|>|: ref(key) 
                                                \/ & \dots \verb|[| accusative \verb|]| )))))) \\
                                            \end{tabular}}\\
                     \end{tabular}\\
\end{tabular}}

(2) Sie wirft den Schl\"ussel weg. (She throws away the key.)

{\footnotesize
\begin{tabular}[t]{@{}l@{}l@{}}
    ( cause ( & {\bf act ( \begin{tabular}[t]{@{}l@{}l@{}}
                          \verb|<|agens, act\verb|>|: ref(she) 
                          \/ & \dots \verb|[| nominative \verb|]| ) \\
                      \end{tabular}} \\ 
              & et ( \begin{tabular}[t]{@{}l}
                         bec ( have ( \begin{tabular}[t]{@{}l@{}l}
                                          \verb|<|goal, have\verb|>|: a1, 
                                          \/ & \dots \verb|[| --- \verb|]|\\ 
                                          \verb|<|to-obj, have\verb|>|: ref(key) 
                                          \/ & \dots \verb|[| --- \verb|]| )) \\
                                           \end{tabular}\\
                         bec ( not ( {\bf have ( \begin{tabular}[t]{@{}l@{}l}
                                                     \verb|<|source, have\verb|>|: ref(she), 
                                                     \/ & \dots \verb|[| --- \verb|]| \\
					             \verb|<|from-obj, have\verb|>|: ref(key) 
                                                     \/ & \dots \verb|[| accusative \verb|]| )))))) \\
                                            \end{tabular}}\\
                     \end{tabular}\\
\end{tabular}}

(3) Er schickt ihm eine Einladung. (He sends him an invitation.)

{\footnotesize
\begin{tabular}[t]{@{}l@{}l@{}}
    ( cause ( & {\bf act ( \begin{tabular}[t]{@{}l@{}l@{}}
                          \verb|<|agens, act\verb|>|: ref(he) 
                          \/ & \dots \verb|[| nominative \verb|]| ) \\
                      \end{tabular}} \\ 
              & et ( \begin{tabular}[t]{@{}l}
                         bec ( {\bf have ( \begin{tabular}[t]{@{}l@{}l}
                                               \verb|<|goal, have\verb|>|: ref(him), 
                                               \/ & \dots \verb|[| dative \verb|]|\\ 
                                               \verb|<|to-obj, have\verb|>|: ref(invitation) 
                                               \/ & \dots \verb|[| accusative \verb|]| )) \\
                                           \end{tabular}}\\
                         bec ( not ( have ( \begin{tabular}[t]{@{}l@{}l}
                                                \verb|<|source, have\verb|>|: ref(he), 
                                                \/ & \dots \verb|[| --- \verb|]| \\
			   	                \verb|<|from-obj, have\verb|>|: ref(invitation) 
                                                \/ & \dots \verb|[| --- \verb|]| )))))) \\
                                            \end{tabular}\\
                     \end{tabular}\\
\end{tabular}}

(4) Er schickt eine Einladung an ihn. (He sends an invitation to him.)

{\footnotesize
\begin{tabular}[t]{@{}l@{}l@{}}
    ( cause ( & {\bf act ( \begin{tabular}[t]{@{}l@{}l@{}}
                          \verb|<|agens, act\verb|>|: ref(he) 
                          \/ & \dots \verb|[| nominative \verb|]| ) \\
                      \end{tabular}} \\ 
              & et ( \begin{tabular}[t]{@{}l}
                         bec ( have ( \begin{tabular}[t]{@{}l@{}l}
                                          \verb|<|goal, have\verb|>|: ref(him), 
                                          \/ & \dots \verb|[| to-phrase \verb|]|\\ 
                                          \verb|<|to-obj, have\verb|>|: ref(invitation) 
                                          \/ & \dots \verb|[| --- \verb|]| )) \\
                                           \end{tabular}\\
                         bec ( not ( {\bf have ( \begin{tabular}[t]{@{}l@{}l}
                                                     \verb|<|source, have\verb|>|: ref(he), 
                                                     \/ & \dots \verb|[| --- \verb|]| \\
					             \verb|<|from-obj, have\verb|>|: ref(invitation) 
                                                     \/ & \dots \verb|[| accusative \verb|]| )))))) \\
                                            \end{tabular}}\\
                     \end{tabular}\\
\end{tabular}}

\caption{Basic semantic forms of sample sentences \label{exs}}
\rule{\textwidth}{0.2mm}
\end{figure*}

For reasons of illustration we have chosen a specific lexical field.
However the principles of the theory of semantic emphasis
also apply to other fields \cite{KunzeFirzlaff93}, e.g.,
change-of-location, 
creation,
measuring, verba dicendi. For each of these fields the theory provides a
prototypical description (i.e. a basic semantic scheme) to which the
rules we presented here and other constraints must be applied in order
to derive a specification of possible grammatical realizations.

\section{\protect{\large \bf GENERATION ARCHITECTURE}}

\label{architecture}

The general architecture of the  {\sc komet} system has been described
in  detail   elsewhere~\cite{Bateman-etal91-penang};  it  follows very
closely  the modularities  entailed by  the  linguistic stratification
assumed    within            systemic-functional           linguistics
(e.g.,~\cite{Halliday78}).  Of most relevance here is the necessity of
specifying the     relationship      between   an    abstract     {\em
grammatically}-oriented semantics  and an account  of the {\em context
of   situation}. This  relationship   underlies the  main   reason for
adopting  a   systemic-functional   orientation   in  text generation:
grammatical  and  lexical decisions  are related to  the deployment of
communicative goals  in   their  communicative context.    The project
includes  the  development of a  large  systemic-functional grammar of
German~\cite{Teich92-gramdoc} and the construction on the basis of the
original Penman English  Upper Model~\cite{UM-89}  of a revised  upper
model ontology that spans both the semantic requirements of German and
English~\cite{Henschel93,HenschelBateman94-coling}.  Input     to  the
grammar component is expressed in  the Penman Sentence Plan Language (
{\sc spl})~\cite{Kasper89-spl}. However,  in contrast to Penman, where
the {\sc spl} is largely equivalent to A-Box assertions made against a
T-Box component combining the  Upper Model and  domain model,  in {\sc
komet} we allocate the generation  system external domain model to the
higher  stratum of context.  This provides  the  theoretical space for
the  flexible  mapping  from  domain  model concepts  to  Upper  Model
concepts   required.      Context   is    organized     into     three
areas~\cite{Halliday78,Martin92} ---  only one  of which, {\em  field}
(the socially  signifiant  activities, participant-types and  activity
sequences of the communicative context), is relevant to us here.

We relate the  information of  the theory  of semantic  emphasis to this
architecture as follows.           We  adopt basic semantic  schemes
 as abstract  general characterizations  of a  subtype of
the field of context; i.e., one  of the contexts in which  interlocutors
can understand  themselves  to  be  is  classifiable  abstractly as, for
example, an exchange of (generalized) possession.  This is then  related
to the  semantic  classes  available  in  the  Upper  Model  by means of
realization:  according to  the distribution  of semantic  emphasis over
the basic semantic scheme, a particular Upper Model concept is selected as
appropriate together with a particular configuration of semantic  roles.
This semantic  specification  then  forms  the  basis  of  possible  {\sc spl}
expressions that can be passed  on to the lexicogrammar  for expression.
As described in~\cite[Chapter 9]{MatthiessenBateman91}, the Upper  Model
is only  one  of  three  bodies  of  semantic  information necessary for
generation. Semantic emphasis distribution also shows co-constraints with
decisions in another component,  the  {\em  Text  Base}, where information concerning
textual statuses  such  as  thematicity, given-/newness,  identifiability,  etc.   is
maintained for  constraining  those  grammatical  decisions sensitive to
such  distinctions.        To  these  we  add  a  component  for   {\em
grammaticalized semantic emphasis}, which represents the distribution of
semantic    emphasis     {\em     that     is     visible    from    the
grammar}.\footnote{\label{gram-emphasis}This  is   entirely   analogous   to
Jackendoff's~\cite[p404/5]{Jackendoff83} view of `argument structure' as
an abbreviation for that part of conceptual structure that is  ``visible
from the syntax'' ---  this is simply  extended systemically to  include
representations of textual  statuses.} In  the examples  below, we  will
represent such textual statuses as additional annotations present in the
{\sc spl}  semantic  specifications;  this  is  the  normal way in which
textual information is captured in {\sc spl}.

\section{\protect{\large \bf EXAMPLES}}

\label{examples}

Given that each semantic form both has a
propositional content ({\em Sachverhaltsrepr\"asentation}) and
indicates, given a particular emphasis distribution, a particular
textual status of the participants in the proposition, we elaborate two
examples of how we can make use of these two aspects in our generation
architecture:
\begin{itemize}
\item {\bf Example 1}: of emphasis information providing grounds on which a 
process type in the
Upper Model can be chosen.
\item {\bf Example 2}: of the mutual constraints between emphasis information and textual
statuses.

\end{itemize}

\noindent {\bf Example 1: Choice of Upper Model type}

Since the introduction of the Penman Upper Model (e.g.,~\cite{UM-89})
in 1985, interfacing with a generation system by means of an abstract
linguistically motivated `ontology' has become widespread (cf., e.g.,
the ontologies of the {\sc lilog} system~\cite{Klose-etal92} and many
others; see~\cite{Bateman92-ontology-tu} for extensive discussion).
Although this is usually achieved by direct {\em subordination} of
domain concepts to `Upper Model' (or equivalent) concepts, this is
known to be insufficient --- domain concepts often need to change
their Upper Model classification depending on their appearance in
particular texts and text organizations.
Here we illustrate how
this general problem of flexibly allocating domain concepts to
appropriate Upper Model concepts can be  partially solved by
an allocation of the semantic emphasis theory.

Our illustration of  the control  of {\em  semantic} choice  by emphasis
information is drawn from the field of {\em change-of-possession}.  This
already constrains  the  possible  choices  of  an  Upper  Model type of
process of a proposition to be  verbalized to {\em
action}.\footnote{By   application    of    a    notion    similar    to
Jackendoff's~\cite[p26]{Jackendoff90} `semantic field feature'.} Without
this specification of field,  choice between all  four process types  in
the Upper Model is completely open.  As shown in  Section~\ref{emphase},
a field specification of change-of-possession has the maximum case frame:
 \begin{math}   <   \end{math}agens,   act\begin{math}   >    \end{math},
\begin{math}   <   \end{math}goal,   have\begin{math}   >    \end{math},
\begin{math}   <   \end{math}to-obj,   have\begin{math}   >   \end{math}
\begin{math}  <  \end{math}source,  have\begin{math}  >  \end{math}  and
\begin{math} < \end{math}from-obj, have\begin{math} > \end{math}.  As an
example, we will consider the cases where two or three of these five
roles have been blocked according to particular emphasis
distributions. In such cases, a process type action with an
instantiation of two Upper Model roles must be chosen. If the
\begin{math}<\end{math}agens,act\begin{math}>\end{math} is blocked and
the \begin{math}<\end{math}source, have\begin{math}>\end{math} and the
\begin{math}<\end{math}from-obj, have\begin{math}>\end{math} have
emphasis and are not blocked or if both the
\begin{math}<\end{math}agens,act\begin{math}>\end{math} and the
\begin{math}<\end{math}from-obj,have\begin{math}>\end{math} have emphasis and are not blocked and the \begin{math}<\end{math}source,have\begin{math}>\end{math} has emphasis and is blocked,
then {\em action process}, subtype {\em dispositive-material-action},
must be chosen (the relevant information is
highlighted):\footnote{This situation corresponds to examples (1) and
(2) in Figure 1. Note also that in the case of the
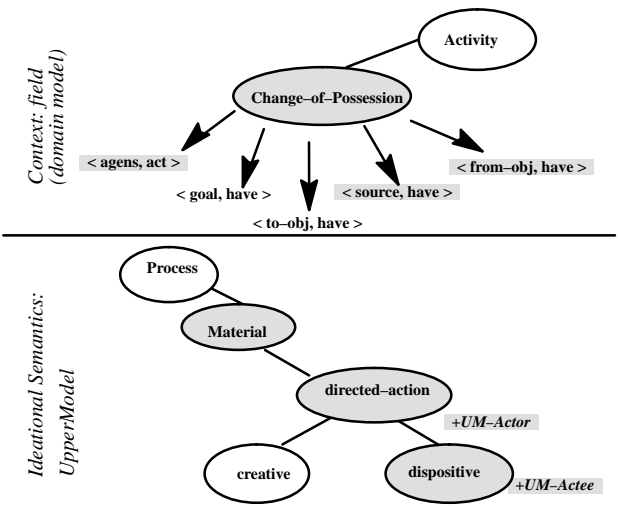
\begin{math}<\end{math}agens,act\begin{math}>\end{math} not being
blocked, it will be co-referential with the
\begin{math}<\end{math}source, have\begin{math}>\end{math}.}

\centerline{\hbox{\psfig{figure=figure.eps}}}






A representation of  a  situation type, such  as change-of-possession,
plus  emphasis information thus makes  it possible to constrain choice
and, in a number of cases,  even determine choice  of a concept in the
Upper Model.

{\bf Example 2: Emphasis distribution and textual status}

For illustration of the control of emphasis
distribution by textual statuses, we have chosen the example of dative shift.   
Dative shift is
motivated by the perspective on a process and the focus/nonfocus on a particular
participant in that  process in a {\em text}, i.e.,  very broadly,  dative shift  is {\em textually} motivated.   
More  specifically,  it  is  motivated  here  by
invoking  a  specific  emphasis  distribution.   We  represent  emphasis
information as  attributed  to  Upper  Model  roles  of  the  {\sc spl}
representation  of  a  clause  in  terms  of  inquiries.\footnote{For  a
description of the  mechanism of  the chooser/inquiry  interface between
semantics and grammar in Penman-type generation architectures, of  which
{\sc komet} is  an example,  see \cite{Mann83-anatomy}.}  Only with  the
emphasis information can we distinguish  between a dative-shifted and  a
nondative-shifted grammatical realization of the ideational part of  the
{\sc spl}:\footnote{These correspond to examples (3) and (4) in Figure 1.}
 
Sample {\sc spl} 1:  Er schickt ihm eine Einladung. (He sends him an invitation.)
{\footnotesize \tt
\begin{tabbing}
\ \ \ (send \=/ directed-action\\
\>	:actor (he / person)\\
\>	:reci\=pient\\
\>\> (him \=/ person\\
\>\>\>  \bf :emphasis-q emphatic)\\
\>	:actee (invitation / object))
\end{tabbing}}
Sample {\sc spl} 2:  Er schickt eine Einladung an ihn. (He sends an invitation to him.)
{\footnotesize \tt
\begin{tabbing}
\ \ \ (send \=/ directed-action\\
\>	:actor (he / person)\\
\>	:reci\=pient\\
\>\> (him \=/ person\\
\>\>\> \bf :emphasis-q nonemphatic)\\
\>	:actee (invitation / object))
\end{tabbing}}

In sample {\sc spl} 1, the recipient is verbalized with emphasis (by
dative case) and the actee is in the focus position.  In sample {\sc
spl} 2, on the other hand, it is the recipient that is in focus
position. Here, it does not have emphasis (and is thus
assigned oblique case).

In our  current architecture~\cite{Bateman-etal93-enlgw}, this kind of
textual   variation is represented  in   a {\em local-level  discourse
semantics}  that  mediates   information   between  the   global-level
discourse organization (represented as  stages in a  generic structure
potential (GSP; cf.  \cite{Hasan78}))  and rhetorical structures (RST;
cf.   \cite{MannThompson85}))   and  the  grammar.   The   local-level
discourse  semantics  (based   on~\cite{Martin92})   contains  textual
linguistic  information   that  controls the  {\em textually-relevant}
options in   the grammar,  such   as topic  and  focus selection  (cf.
\cite{Sgall-etal86}), reference  and information  structure.  Given  a
representation of propositional  content, the text planner keeps track
of textual decisions in thematic development and reference attribution
and selects from  the {\em textually-relevant} emphasis potential that
option  that is appropriate  in a given context.   Consider a piece of
text  as   it  typically  occurs  in  the   domain   we deal  with  in
\cite{Teich-etal94-anlp} (arts and artists' biographies) that provides
a context for the choice of  emphasis distribution in sample {\sc spl}
1:

\noindent
{\em   (1)  Seit  1898     besch\"aftigte    sich Behrens  mit     den
Gestaltungsproblemen   von  Industrieprodukten. (2)  Er entwarf  unter
anderem  Flaschen  f\"ur  die    Serienherstellung in einer    grossen
Glasfabrik.  (3) 1899} schickte  der  Grossherzog von Hessen ihm  eine
Einladung {\em  nach Darmstadt zu kommen und  sich einer Gruppe junger
K\"unstler anzuschliessen ...}\footnote{ English: {\em In 1898 Behrens
turned to problems  of industrial production and  designed a number of
prototype flasks for mass production by a large  glass works. In 1899}
the Grand Duke  of  Hessen sent him  an  invitation  {\em to come   to
Darmstadt and join a group of young artists...}}

Typically, in biography texts, the artist the text deals with acts as
the {\em hypertheme} of the text. Moreover, one of the typical
thematic developments by which a biography text proceeds is selecting
temporal locations (as in (1)) or reselecting the hypertheme as theme
of the next sentence (as in (2)).  This textual organization is
accordingly produced by the text planner. Then, given this textual
status, all references to the participant constituting the hypertheme
(here: {\em Behrens}) belong to information already introduced. Being
the {\em hypertheme} and the {\em given information} is reflected in
the participant receiving emphasis status. Grammatically, this is
realized in the assignment of nonoblique case to the participant and
its ordering in the clause.  In sentence (3), the recipient role thus
receives emphasis status and cannot appear in the focus position
which is generally reserved for pieces of information that are new in
the discourse and not thematic (cf. * {\em Er schickte eine Einladung ihm}). 
The problematic gap between the high-level
textual organization and grammatical expression is thus appropriately
bridged.

\section{\protect{\large \bf CONCLUSIONS, SIGNIFICANCE AND FUTURE WORK}}

\label{future}

We have shown that emphasis information can provide more control of
choices in generation on the higher strata of the linguistic system
(semantics).  {\em Ideationally}, emphasis distribution and blocking
of roles constrains the possible process types of the Upper Model to
be chosen.  In the grammar, it consequently constrains choice in case
assignment.
We have also sketched the aspect of emphasis theory that is relevant
for {\em textual} decisions in thematicity and information structure which
attribute certain textual statuses to the participants in the discourse. These
are reflected grammatically for example in relation changing phenomena such as dative shift.
In a current application of NL analysis and
generation components to the domain of arts and artists' biographies
\cite{Rostek-etal94}, the mechanisms described above in the discussion of 
example 1 provide one component of a domain model that is used both by analysis
and generation \cite{Teich-etal94-anlp}.

Some next steps for this work are clear.  Many of the examples put
forward by Kunze are argued in terms of textual acceptability that
goes beyond single clauses.  We are now, therefore, investigating the
relationship of emphasis information and textual statuses we have
sketched in the discussion of example 2 more closely, also considering
other fields, such as creation, change-of-location and verba dicendi.

A  further  step  is  to  investigate the  multilingual
applicability of the framework --- for example, in~\cite{Kunze92}, Kunze
proposes an analogous treatment for the field of change-of-possession  in
English. It will be  interesting to investigate the applicability to
English of a detailed  account that has  been worked on  the basis of  a
language other than  English ---  the reverse  of what  normally occurs!
Semantic emphasis  may  support  an  improved  interface between textual
organization and grammatical  decisions for  English also,  although, at
least in  a  systemic-functional  account,  somewhat  more  functionally
differentiated proposals have  been made  for the  phenomena that  Kunze
gathers  together  under  semantic  emphasis  (for  example,   given-new
information,  theme-rheme  information,  and  modal  responsibility   of
the grammatical subject ---  all of  which are  independently variable; cf. \cite{Martin92}). The precise
relationship of semantic emphasis to these needs to be clarified.

 
\bibliographystyle{named}

{\small
\begin{thebibliography}{}

\bibitem[\protect\citeauthoryear{Alshawi}{1992}]{Alshawi92}
Hiyan Alshawi, editor.
\newblock {\em The Core Language Engine}.
\newblock MIT Press, Cambridge, Massachusetts, 1992.

\bibitem[\protect\citeauthoryear{Bateman \bgroup \em et al.\egroup
  }{1990}]{UM-89}
John~A. Bateman, Robert~T. Kasper, Johanna~D. Moore, and Richard~A. Whitney.
\newblock A general organization of knowledge for natural language processing:
  the {\sc penman} upper model.
\newblock Technical report, USC/Information Sciences Institute, Marina del Rey,
  California, 1990.

\bibitem[\protect\citeauthoryear{Bateman \bgroup \em et al.\egroup
  }{1991}]{Bateman-etal91-penang}
John~A. Bateman, Elisabeth~A. Maier, Elke Teich, and Leo Wanner.
\newblock Towards an architecture for situated text generation.
\newblock In {\em International Conference on Current Issues in Computational
  Linguistics}, Penang, Malaysia, 1991.
\newblock Also available as technical report of GMD/Institut f{\"u}r
  Integrierte Publikations- und Informationssysteme, Darmstadt, Germany.

\bibitem[\protect\citeauthoryear{Bateman \bgroup \em et al.\egroup
  }{1993}]{Bateman-etal93-enlgw}
John~A. Bateman, Liesbeth Degand, and Elke Teich.
\newblock Multilingual textuality: Some experiences from multilingual text
  generation.
\newblock In {\em Proceedings of the Fourth European Workshop on Natural
  Language Generation, Pisa, Italy, 28-30 April 1993}, pages 5 -- 17, 1993.
\newblock Also available as technical report from GMD/Institut f{\"u}r
  Integrierte Publikations- und Informationssysteme, Darmstadt, Germany.

\bibitem[\protect\citeauthoryear{Bateman}{1992}]{Bateman92-ontology-tu}
John~A. Bateman.
\newblock The theoretical status of ontologies in natural language processing.
\newblock In Susanne Preu\ss\ and Birte Schmitz, editors, {\em Text
  Representation and Domain Modelling -- ideas from linguistics and {AI}},
  pages 50 -- 99. KIT-Report 97, Technische Universit{\"at} Berlin, May 1992.
\newblock (Papers from KIT-FAST Workshop, Technical University Berlin, October
  9th - 11th 1991).

\bibitem[\protect\citeauthoryear{Fillmore}{1968}]{Fillmore68}
Charles~J. Fillmore.
\newblock The case for case.
\newblock In Emons Bach and Robert~T. Harms, editors, {\em Universals in
  Linguistic Theory}. Holt, Rinehart and Wilson, New York, 1968.

\bibitem[\protect\citeauthoryear{Firzlaff and
  Haenelt}{1992}]{FirzlaffHaenelt92}
Beate Firzlaff and Karin Haenelt.
\newblock On the acquisition of conceptual definitions via textual modelling of
  meaning paraphrases.
\newblock In {\em Proceedings of the fifteenth International Conference on
  Computational Linguistics (COLING-92)}, volume~{IV}, pages 1209 -- 1213.
  International Committe on Computational Linguistics, 1992.

\bibitem[\protect\citeauthoryear{Grosz and Sidner}{1986}]{GroszSidner86}
Barbara~J. Grosz and Candace~L. Sidner.
\newblock Attention, intentions and the structure of discourse.
\newblock {\em Computational Linguistics Journal}, 12(3):175--204, 1986.

\bibitem[\protect\citeauthoryear{Grover \bgroup \em et al.\egroup
  }{1993}]{Grover-etal93-alvey}
Claire Grover, John Carroll, and Ted Briscoe.
\newblock {The Alvey Natural Language Tools Grammar (4th release)}.
\newblock Technical report, Human Communication Research Centre, University of
  Edinburgh and Computer Laboratory, University of Cambridge, 1993.

\bibitem[\protect\citeauthoryear{Halliday}{1978}]{Halliday78}
Michael~A.K. Halliday.
\newblock {\em Language as social semiotic}.
\newblock Edward Arnold, London, 1978.

\bibitem[\protect\citeauthoryear{Hasan}{1978}]{Hasan78}
Ruqaiya Hasan.
\newblock {Text in the Systemic-Functional Model}.
\newblock In Wolfgang Dressler, editor, {\em Current Trends in Text
  Linguistics}, pages 228--246. de Gruyter, Berlin, 1978.

\bibitem[\protect\citeauthoryear{Henschel and
  Bateman}{1994}]{HenschelBateman94-coling}
Renate Henschel and John Bateman.
\newblock The merged upper model: a linguistic ontology for {G}erman and
  {E}nglish.
\newblock In {\em Proceedings of {COLING} '94}, Kyoto, Japan, August 1994.

\bibitem[\protect\citeauthoryear{Henschel}{1993}]{Henschel93}
Renate Henschel.
\newblock {Merging the English and the German Upper Model}.
\newblock Technical report, GMD/Institut f{\"u}r Integrierte Publikations- und
  Informationssysteme, Darmstadt, Germany, 1993.

\bibitem[\protect\citeauthoryear{Horacek and Zock}{1993}]{horacekzock93}
Helmut Horacek and Michael Zock, editors.
\newblock {\em New concepts in natural language generation}.
\newblock Pinter Publishers, London, 1993.

\bibitem[\protect\citeauthoryear{Hovy \bgroup \em et al.\egroup
  }{1992}]{Hovy-etal92-nlgw}
Eduard Hovy, Julia Lavid, Elisabeth Maier, Vibhu Mittal, and Cecile Paris.
\newblock Employing knowledge resources in a new text planner architecture.
\newblock In {\em Proceedings of the 6th International Workshop on Natural
  Language Generation}, Trento, Italy, 1992. Springer-Verlag.

\bibitem[\protect\citeauthoryear{Hovy}{1987}]{Hovy87-thesis}
Eduard~H. Hovy.
\newblock {\em Generating Natural Language under Pragmatic Constraints}.
\newblock PhD thesis, Yale University, 1987.
\newblock (Technical report: YALEU/CSD/RR 521). Also published by Lawrence
  Erlbaum Associates, Hillsdale, New Jersey, 1988.

\bibitem[\protect\citeauthoryear{Jackendoff}{1983}]{Jackendoff83}
Ray Jackendoff.
\newblock {\em Semantics and Cognition}.
\newblock The M.I.T. Press, Cambridge, MA, 1983.

\bibitem[\protect\citeauthoryear{Jackendoff}{1990}]{Jackendoff90}
Ray Jackendoff.
\newblock {\em Semantic Structures}.
\newblock The M.I.T. Press, Cambridge, MA, 1990.

\bibitem[\protect\citeauthoryear{Kasper}{1989}]{Kasper89-spl}
Robert~T. Kasper.
\newblock A flexible interface for linking applications to {\sc penman}'s
  sentence generator.
\newblock In {\em Proceedings of the {DARPA} Workshop on Speech and Natural
  Language}, 1989.
\newblock Available from USC/Information Sciences Institute, Marina del Rey,
  CA.

\bibitem[\protect\citeauthoryear{Klose \bgroup \em et al.\egroup
  }{1992}]{Klose-etal92}
Gudrun Klose, Ewald Lang, and Thomas Pirlein, editors.
\newblock {\em Die {O}ntologie und {A}xiomatik der {W}issensbasis von {LILOG}:
  {Wissensmodellierung im IBM Deutschland LILOG-Projekt}}.
\newblock Springer-Verlag, Berlin/Heidelberg, 1992.
\newblock Informatik-Fachberichte 307.

\bibitem[\protect\citeauthoryear{Kunze and Firzlaff}{1993}]{KunzeFirzlaff93}
J{\"u}rgen Kunze and Beate Firzlaff.
\newblock {\em {Sememstrukturen und Feldstrukturen}}, volume XXXVI of {\em
  Studia Grammatica}.
\newblock Akademie Verlag, Berlin, 1993.

\bibitem[\protect\citeauthoryear{Kunze}{1991}]{Kunze91}
J{\"u}rgen Kunze.
\newblock {\em Kasusrelationen und semantische Emphase}, volume XXXII of {\em
  Studia Grammatica}.
\newblock Akademie Verlag, Berlin, 1991.

\bibitem[\protect\citeauthoryear{Kunze}{1992}]{Kunze92}
J{\"ur}gen Kunze.
\newblock {Verbfeldstrukturen und {\"Ubersetzung}}.
\newblock {\em {Zeitschrift f{\"u}r Literaturwissenschaft und Linguistik}},
  84:67 -- 103, 1992.

\bibitem[\protect\citeauthoryear{Lascarides and
  Asher}{1991}]{LascaridesAsher91}
Alex Lascarides and N.~Asher.
\newblock Discourse relations and defeasible knowledge.
\newblock In {\em Proceedings of the 29th. Annual Meeting of the Association
  for Computational Linguistics}, pages 55 -- 63, Berkeley, California, 1991.
  Association for Computational Linguistics.

\bibitem[\protect\citeauthoryear{Mann and Thompson}{1985}]{MannThompson85}
William~C. Mann and Sandra~A. Thompson.
\newblock Assertions from discourse structure.
\newblock In {\em Proceedings of the Eleventh Annual Meeting of the Berkeley
  Linguistics Society}, Berkeley, 1985. Berkeley Linguistic Society.
\newblock Also available as USC/Information Sciences Institute, Technical
  Report ISI/RS-85-155.

\bibitem[\protect\citeauthoryear{Mann and Thompson}{1987}]{MannThompson87}
William~C. Mann and Sandra~A. Thompson.
\newblock Rhetorical structure theory: a theory of text organization.
\newblock Technical Report RS-87-190, USC/Information Sciences Institute, 1987.
\newblock Reprint series.

\bibitem[\protect\citeauthoryear{Mann}{1983}]{Mann83-anatomy}
William~C. Mann.
\newblock The anatomy of a systemic choice.
\newblock {\em Discourse Processes}, 1983.
\newblock Also available as USC/Information Sciences Institute, Research Report
  ISI/RR-82-104, 1982.

\bibitem[\protect\citeauthoryear{Martin}{1992}]{Martin92}
James~R. Martin.
\newblock {\em English text: systems and structure}.
\newblock Benjamins, Amsterdam, 1992.

\bibitem[\protect\citeauthoryear{Matthiessen and
  Bateman}{1991}]{MatthiessenBateman91}
Christian~M.I.M. Matthiessen and John~A. Bateman.
\newblock {\em Text generation and systemic-functional linguistics: experiences
  from {E}nglish and {J}apanese}.
\newblock Frances Pinter Publishers and St. Martin's Press, London and New
  York, 1991.

\bibitem[\protect\citeauthoryear{Matthiessen \bgroup \em et al.\egroup
  }{1991}]{Matthiessen-etal91-kyushu}
Christian~M.I.M. Matthiessen, Michael O'Donnell, and Licheng Zeng.
\newblock Discourse analysis and the need for functionally complex grammar in
  parsing.
\newblock In {\em Proceedings of the 2nd Japan-Australia Joint Symposium on
  Natural Language Processing}, Kyushu, Japan, 1991. Kyushu Institute of
  Technology.

\bibitem[\protect\citeauthoryear{Meteer}{1991}]{Meteer91}
Marie~W. Meteer.
\newblock Bridging the generation gap between text planning and linguistic
  realization.
\newblock {\em Computational Intelligence}, 7(4):296 -- 304, 1991.

\bibitem[\protect\citeauthoryear{Rostek \bgroup \em et al.\egroup
  }{1994}]{Rostek-etal94}
Lothar Rostek, Wiebke M\protect{\"o}hr, and Dietrich~H. Fischer.
\newblock Weaving a web: The structure and creation of an object network
  representing an electronic reference network.
\newblock In {\em Proceedings of Electronic Publishing (EP) '94}, 1994.

\bibitem[\protect\citeauthoryear{Sgall \bgroup \em et al.\egroup
  }{1986}]{Sgall-etal86}
Petr Sgall, Eva Haji\v{c}ov\'a, and J.~Panevov\'a.
\newblock {\em The Meaning of the Sentence in Its Semantic and Pragmatic
  Aspects}.
\newblock Reidel Publishing Company, Dordrecht, 1986.

\bibitem[\protect\citeauthoryear{Teich \bgroup \em et al.\egroup
  }{1994}]{Teich-etal94-anlp}
Elke Teich, Beate Firzlaff, and Lothar Rostek.
\newblock Using {NLP} in a system supporting electronic publishing.
\newblock Technical report, Institut f{\"u}r Integrierte Publikations- und
  Informationssysteme (IPSI), GMD, Darmstadt, 1994.

\bibitem[\protect\citeauthoryear{Teich}{1992}]{Teich92-gramdoc}
Elke Teich.
\newblock Komet: grammar documentation.
\newblock Technical report, GMD/Institut f{\"ur} Integrierte Publikations- und
  Informationssysteme, Darmstadt, West Germany, 1992.

\end{thebibliography}}

\end{document}